\pdfoutput=1

\documentclass[11pt]{article}

\usepackage[acceptedWithA]{tacl2021v1}
\usepackage{pgf}
\usepackage{times}
\usepackage{latexsym}

\usepackage{tikz} 
\usepackage[T1]{fontenc}

\usepackage[utf8]{inputenc}

\usepackage{microtype}

\usepackage{inconsolata}

\usepackage{graphicx}
\usepackage{amsmath}
\usepackage{bbm}

\usepackage[capitalize]{cleveref}
\crefname{section}{Sec.}{Secs.}
\Crefname{section}{Section}{Sections}
\crefname{table}{Table.}{Tables.}
\Crefname{table}{Table}{Tables}
\usepackage{color, colortbl}
\definecolor{Gray}{gray}{0.93}
\usepackage{multirow}
\usepackage{svg}
\usepackage{amssymb}

\usepackage{threeparttable}
\usepackage{multirow}
\usepackage{booktabs}
\usepackage{adjustbox}

\usepackage{colortbl}
\usepackage{array}
\usepackage{pifont}

\usepackage[table]{xcolor}   
\usepackage{pgf}            
\usepackage{siunitx}        
\usepackage{ifthen} 
\usepackage{fontawesome5}

\newcommand{\datasetlong}{ASL Minimal Translation Pairs}
\newcommand{\dataset}{\textsc{ASL-MTP}}
\newcommand{\datasetshort}{\textsc{ASL-MTP}}
\newcommand{\shubert}{\textsc{SHuBERT}}
\newcommand{\asllrp}{\textsc{asllrp}}
\newcommand{\bleurt}{BLEURT}

\newcommand{\heatmap}[1]{%
  \pgfmathsetmacro{\maxpos}{50}  
  \pgfmathsetmacro{\maxneg}{20}  
  \pgfmathsetmacro{\v}{{#1}}                 
  \pgfmathsetmacro{\vabs}{abs(\v)}         
  \pgfmathparse{\v<0}\ifnum\pgfmathresult=1
      \pgfmathsetmacro{\ratio}{min(1,\vabs/\maxneg)}   
      \pgfmathsetmacro{\percent}{\ratio*100}           
      \cellcolor{green!\percent!white}%
  \else
      \pgfmathsetmacro{\ratio}{min(1,\v/\maxpos)}      
      \pgfmathsetmacro{\percent}{\ratio*100}           
      \cellcolor{red!\percent!white}%
  \fi
  \num[round-mode=places,round-precision=2]{#1}%
}

\definecolor{greeen}{HTML}{66a61e}
\definecolor{reed}{HTML}{b2182b}

\title{Targeted Linguistic Analysis of Sign Language Models\\with Minimal Translation Pairs}

\author{
 \textbf{Serpil Karabüklü},\textsuperscript{1}
 \textbf{Kanishka Misra},\textsuperscript{1,2,*}
 \textbf{Shester Gueuwou},\textsuperscript{1}\\
 \textbf{Diane Brentari,\textsuperscript{3}}
 \textbf{Greg Shakhnarovich,\textsuperscript{1}}
 \textbf{Karen Livescu\textsuperscript{1}}
\\
\\
\small { \textsuperscript{1}Toyota Technological Institute at Chicago,
 \textsuperscript{2}Linguistics Department, The University of Texas at Austin,}\\
 \small{\textsuperscript{3}Linguistics Department, The University of Chicago}\\
 \small{
    \texttt{\{skarabuklu, shesterg, greg, klivescu\}@ttic.edu, kmisra@utexas.edu, dbrentari@uchicago.edu}
 }
}

\begin{document}
\maketitle
\thispagestyle{plain}

{\let\thefootnote\relax\footnotetext{$^{*}$Work done partly while at TTIC}}

\begin{abstract}
Models of sign language have historically lagged behind those for spoken language (text and speech). Recent work has greatly improved their performance on tasks like sign language translation and isolated sign recognition. However, it remains unclear to what extent existing models capture various linguistic phenomena of sign language, and how well they use cues from the multiple articulators used in sign language (hands, upper body, face). We introduce a new benchmark dataset for American Sign Language, \datasetlong{} (\dataset), divided into multiple types of sign language phenomena and corresponding minimal pairs of translations, for performing such linguistic analyses. As a case study, we use \dataset{} to analyze a state-of-the-art ASL-to-English translation model. We conduct a targeted analysis of the model by ablating various input cues during training and inference and evaluating on the phenomena in \dataset{}. Our results show that, while the model performs above chance level on most of the phenomena, it relies strongly on manual cues while often missing crucial non-manual cues.

\vspace{0.5em}
\noindent\faGithub\ \href{https://github.com/serpilkarabuklu/SL-Models-Analysis}{https://github.com/serpilkarabuklu/SL-Models-Analysis}

\end{abstract}  
\section{Introduction}
\label{sec:intro}

\begin{figure}[!t]
    \centering
    \resizebox{\linewidth}{!}{%
        \includegraphics{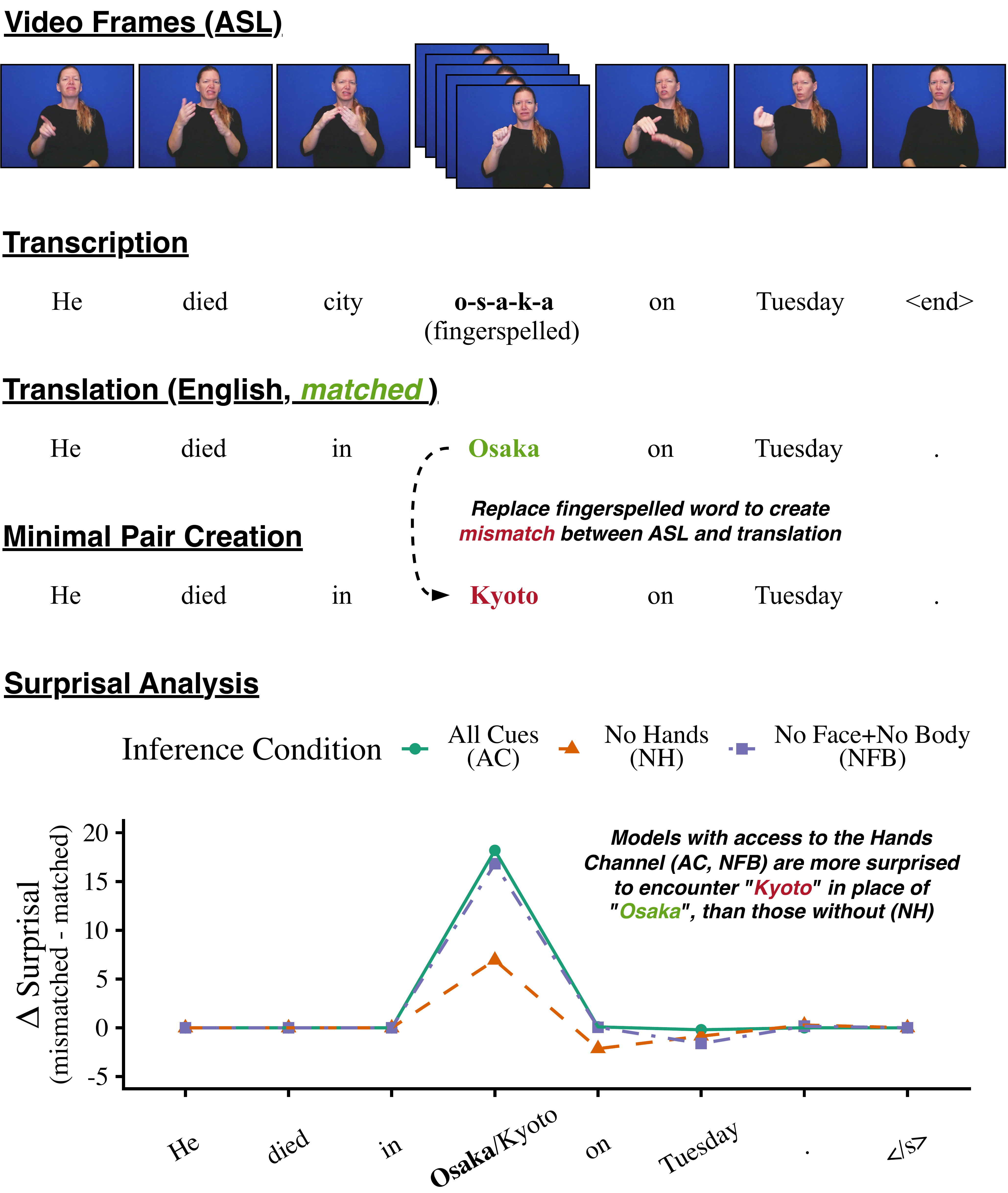}%
    }

    \caption{Our dataset construction and analysis approach. We create minimal pairs for English translations of ASL inputs by replacing critical segments (here, the fingerspelled word, ``Osaka''). We then measure the difference in a sign language translation model's 
    surprisal on the matched vs.~mismatched sequences to quantify the model's sensitivity to the target phenomenon (fingerspelling). The stimulus shown is taken from FLEURS-ASL \citep{tanzer2024fleurs}, and is used only for illustration.}
    \label{fig:nll_sample}
    \vspace{-1em}
\end{figure}

Sign languages convey meaning visually through the combination of multiple articulatory channels, traditionally grouped by linguists into manuals (hands) and non-manuals (facial and body movements)~\citep{valli2000linguistics, sandler2006sign, slhandbook_hsk, quer2021routledge}. Computational models of sign language video have been developed for a variety of tasks, such as isolated sign recognition~\cite{li2020word}, continuous sign recognition (i.e.~glossing)~\cite{camgoz2020sign}, and translation from sign language to written text in a spoken language~\cite{camgoz2018neural, shi2022open, zhang2024scaling}. Recent work has also developed pre-trained representation models for sign language video, in order to enable quick fine-tuning for multiple downstream tasks~\cite{hu2021signbert,gueuwou2024shubertselfsupervisedsignlanguage,wong2025signrep}.
Some models of sign language have focused only on manual signs~\citep{santhalingam2020finehand, hu2021signbert, hu2023signbert+}, some use the entire input images~\citep{shi2022open,rust-etal-2024-towards}, and some divide the input into multiple channels, for both perception~\citep{camgoz2020multi, gueuwou2024signmusketeers,gueuwou2024shubertselfsupervisedsignlanguage} and generation \cite{saunders2020adversarial, ma-etal-2024-multi}.

Recent work has produced substantial improvements in the performance of sign language translation models. However, it is unclear how well such models handle specific linguistic phenomena, and in particular, phenomena that critically depend on channel-specific cues. For instance, phenomena such as Wh- and Polar Questions often involve a combination of hand and eyebrow movements \citep{baker1983microanalysis}, while fingerspelling and numbers largely depend on manual cues. Even for models that explicitly learn from multiple channels, it is unclear if they extract information from these channels in a way that matches expectations from sign language linguistics.

We aim to address the abovementioned gaps, using American Sign Language (ASL) data. We first present \datasetlong{} (\dataset{}), a collection of 
ASL utterance videos along with corresponding (matched, mismatched) pairs of English translation sentences that minimally differ with respect to a particular target phenomenon. Here, a model should assign higher probability to the correct (matched) reference translation than to its minimally differing (mismatched) counterpart.  \datasetshort{} covers 9 phenomena and is inspired by the now well-established practice of using minimal pair datasets to evaluate the linguistic knowledge of language models \citep[][i.a.]{linzen2016assessing, marvin2018targeted, warstadt-etal-2020-blimp-benchmark, hu2025stringprobabilitytellgrammaticality}, as well as contrast sets in machine translation \citep{sennrich-2017-grammatical}. \dataset{} is the first minimal-pair dataset that focuses on phenomena-specific sensitivity in sign language translation models. 

As a case study, we apply minimal pair analysis with \dataset{} to SHuBERT+ByT5~\cite{gueuwou2024shubertselfsupervisedsignlanguage}, a state-of-the-art sign language translation model that uses multiple input channels. This property of the model allows us to manipulate information represented in each channel, to shed light on whether the model extracts information in a manner that conforms to linguistic expectations. Specifically, we devise a set of cue ablations where one or more channels are masked from the input, and test whether doing so affects the model's performance on \dataset{}, especially on phenomena that rely upon the ablated cue.
We find that when the model has access to all cues, it performs above chance on 8 of the 9 phenomena in \datasetshort{}. For one phenomenon, Polar Questions, the model has a strong bias toward declarative sentences over questions. 
When ablating cues, we find mixed results in terms of the effect of ablation on model performance. While the model is clearly affected by the lack of hands, it is not always sensitive to losses in non-manual cues.  We also ablate cues at training time, inspired by controlled rearing of language models~\citep[e.g.,][]{misra-mahowald-2024-language}. 

Our work contributes to finer-grained linguistic evaluation of sign language translation models, and our case study points to a need to improve the use of non-manual cues in a state-of-the-art model.


 
\section{Related Work}
\label{sec:related_work}

\subsection{Sign Language Models}

Sign language processing has traditionally been fragmented into task-specific methods and models, with different architectures and designs. A primary bottleneck limiting unification has been the scarcity of large datasets for pretraining. However, with the emergence of larger datasets such as YouTube-ASL \citep{uthus2023youtube} ($\sim$1,000 hours), recent years have seen the rise of pretrained sign language models that can be adapted with minimal fine-tuning for a variety of downstream tasks.
Both supervised \citep{uthus2023youtube, zhang2024scaling} and self-supervised approaches \citep{rust-etal-2024-towards, wong2025signrep} have advanced considerably, but most large-scale ASL models remain inaccessible as their weights are not publicly released. The only state-of-the-art ASL model of which we are aware that is publicly available and trained on substantial ASL data is \shubert{} \citep{gueuwou2024shubertselfsupervisedsignlanguage}, which we use for our case study here. \shubert{} is a self-supervised ASL video representation model, trained via masked prediction of automatically learned discrete ``tokens'' in multiple channels (hands, face, upper body pose), and has been fine-tuned to obtain state-of-the-art performance on multiple tasks (translation, isolated sign recognition, fingerspelling detection). \shubert{} is described in more detail in~\cref{sec:model_studied}.

Evaluation of sign language models has typically relied on high-level, task-specific measures such as BLEU \citep{papineni2002bleu} and BLEURT \citep{sellam2020bleurt} for translation \citep{camgoz2018neural, shi2022open} and accuracy for isolated sign recognition \citep{kezar2023sem,li2020word,neidle-etal-2022-resources}. Some studies have focused on isolated signs at the phonological level \cite{tornay2020phonology, sandoval2023self, kezar2025american, kezar2025phonologicalrepresentationlearningisolated}, while others on continuous signing have evaluated models on individual linguistic phenomena, such as intensification \cite{inan-etal-2022-modeling}, phonetic reduction driven by discourse effects \cite{imai2025pragmaticsshapearticulationcomputational}, and co-reference resolution of indexical signs \cite{yin-etal-2021-signed}. 
Task-specific measures often miss out on linguistic nuances of sign language, while the existing targeted analyses are limited to a single channel (the hands) or a constrained setting (e.g., isolated signs).

\subsection{Linguistic Analysis of Language Models}
\label{sec:eval-lit}

In contrast with standard task-oriented benchmarks, linguistically motivated analyses of language models (LMs) typically involve a controlled approach to collecting data that isolates a phenomenon of interest \citep{linzen2016assessing, hawkins-etal-2020-investigating, weissweiler-etal-2022-better, wilcox2024using, misra2024generating}. A standard approach to evaluation in this space is the minimal pairs paradigm \citep[][etc.]{warstadt-etal-2020-blimp-benchmark, hu2025stringprobabilitytellgrammaticality}, in which the LM is provided with pairs of sentences that differ in ways that are critical to the phenomenon of interest, where one of them is acceptable and the other unacceptable. As an example of a pair that can be used to evaluate an LM on number agreement, consider ``\textit{The woman \textbf{laughs}.}'' vs.~``\textit{*The woman \textbf{laugh}.}'', which differ only in the number agreement of the verb (\textit{laugh}) with the subject (\textit{woman}). The LM is evaluated on a number of such pairs, by comparing its ``score'' (usually, log-probability per token) on each pair. Accuracy, then, is the proportion of time the LM's score for the acceptable sentence is greater than for the unacceptable sentence.  

There has been a concerted effort to build controlled minimal-pair benchmarks for languages beyond written English \citep{xiang-etal-2021-climp, someya-oseki-2023-jblimp, taktasheva-etal-2024-rublimp, suijkerbuijk2025blimp, jumelet2025multiblimp}. The idea of {\it translation} minimal pairs has been used to study machine translation models' handling of specific linguistic phenomena like agreement and polarity~\cite{sennrich-2017-grammatical}. \datasetshort{} takes inspiration from these and expands the tradition of minimal pair analysis to sign languages.

Recent studies have also analyzed linguistic behavior in LMs by performing controlled training data ablation \citep{jumelet-etal-2021-language, misra-mahowald-2024-language, patil-etal-2024-filtered, leong2024testing, yao2025both, xu2026crossmodaltaxonomicgeneralizationvision}. In these studies, targeted parts of an LM's training corpus are removed or ``ablated'' to test whether models can recover this knowledge from other parts of the corpus. In our case study, we take (loose) inspiration from this approach and ablate specific channels in the model (hands, body, face).

\subsection{Manual and non-manual channels}
\label{sec:manual_nonmanual}

Our work targets phenomena encoded across multiple channels, including the hands, face, and body. Signs encoded in the hands are referred to as manual signs while facial and body movements that convey grammatical functions are referred to as non-manuals \cite[i.a.]{valli2000linguistics, sandler2006sign, karabuklu2024prosody}. For some phenomena, such as Fingerspelling \cite{brentari2001native, Keane2016ac} or Classifiers \cite{benedicto2004did, zwitserlood2012classifiers}, the primary cues are only in the hands (manual signs); others, such as Wh-Questions \cite{baker1983microanalysis, neidle2000syntax}, Negation \cite{veinberg, neidle2000syntax}, and Conditionals \cite{baker-padden, liddell1980american, liddell1986headthrust, wilbur1999syntactic, wilbur2011nonmanuals}, have primary cues which are both manual signs and non-manuals. For instance, Conditionals are conveyed with both the manual sign \textsc{if} and the non-manual eyebrow raise. Finally, some sign language phenomena are solely encoded through non-manuals. For instance, a Polar Question (\textit{Are you ready?}) differs from its declarative counterpart (\textit{You are ready.}) only in terms of eyebrow raise \cite{baker1983microanalysis, weast2008questions}. 

In general, manuals are considered to convey the bulk of the content in sign language~\citep{brentari2019sign}, but other cues also contribute substantially, either as primary or secondary cues~\cite{malaia2018information,benitez2014discriminant}.  In fact, non-manual cues are prevalent across linguistic domains \cite{pfau2010nonmanuals, wilbur2021non}:  phonology \cite{wilbur1994eyeblinks}, morphology \cite{anderson1998pah}, syntax \cite{liddell1980american, liddell1986headthrust, neidle2000syntax, watson2010contentquestion}, semantics and pragmatics \cite{coulter1978raised, shaffer2004information, herrmann2013modal, karabuklu2024simultaneity}. We therefore expect successful sign language models to use information from both manual and non-manual channels.

\normalsize

\definecolor{greeen}{HTML}{66a61e}
\definecolor{reed}{HTML}{b2182b}

\section{\datasetlong{}}
\label{sec:dataset}

One of the main contributions of this work is \datasetlong{} (\dataset{}), a dataset to evaluate fine-grained linguistic capacities of models that translate ASL to English.
\dataset{} consists of 1,275 ASL videos along with corresponding pairs of acceptable and unacceptable written English translations. Since the acceptability of a sentence depends on the extent to which it matches the ASL video, we call the acceptable sentences ``matched'' and unacceptable ones ``mismatched''. The dataset is divided into 9 subsets, each of which targets a specific phenomenon.

The ASL videos and their corresponding sentences were drawn from \asllrp{} \citep{neidle2022aslvideocorpora}, a collection of 2,048 high-quality, linguistically annotated ASL utterances, along with their English translations, produced by 4 signers. \asllrp{} includes annotations for manual signs (e.g., number of hands, hand movements) and time-aligned non-manuals (e.g., head position and movements, mouth movements, eye gaze) along with their grammatical functions (e.g., classifier, question, conditional).  These annotations, combined with the fact that \asllrp{} has not been widely used in the training of sign language models, make it an ideal source for evaluating models.  Although the dataset is not large, it provides enough data to evaluate models on the phenomena of interest and, as we will see, to obtain statistically significant results in our analyses (\cref{sec:case_study}).
Below we provide more details about our phenomenon selection criteria and dataset construction, as well as the intended usage of \dataset{} for minimal pair evaluation.

\subsection{Details of \dataset{} construction}

\paragraph{Phenomena}
To investigate whether sign language models rely on cues from multiple input channels, we selected 9 phenomena that involve a range of channel combinations. Our phenomena can be grouped into three subsets: 1) ones that are mainly encoded in the hands---Numbers, Fingerspelling, and Classifiers; 2) ones that are encoded in both the hands and face---Negation, Wh-Questions, and Conditionals; and 3) ones that are predominantly encoded in the face---Polar Questions.
This grouping is not perfectly clean, because of the existence of varying secondary cues---e.g., there are several stimuli in our dataset where `Conditionals' are signed using non-manual cues. Therefore, we will discuss results on such exceptional cases separately, when relevant. 

\begin{table*}[!ht]
\centering

\footnotesize{

\begin{tabular}{p{2cm}p{4cm}p{4.8cm}p{3.5cm}}

\toprule
\textbf{Phenomenon} & \textbf{Description} & \textbf{Construction} & \textbf{Minimal Pair Examples} \\
\midrule

Numbers ($N$=119)  & Manual signs with phonetically complex handshapes and movements. & \begin{tabular}[t]{@{}p{4.8cm}@{}}\textbf{Filtering Criteria:} Sentences containing numerical values.\\\textbf{Manipulation:} Replace number with a different number.\end{tabular}
 &  \begin{tabular}[t]{@{}p{3.5cm}@{}}\textbf{Matched}: The movie starts at \textcolor{greeen}{\textit{7}}.\\\textbf{Mismatched}: The movie starts at \textcolor{reed}{\textit{8}}.\end{tabular} 
 \\ 
\midrule

Fingerspelling ($N$=170) & Sequences of letter representations of spoken words, signed using handshapes. Mostly used for proper nouns, technical terms, and borrowings without established signs.  & \begin{tabular}[t]{@{}p{4.8cm}@{}} \textbf{Filtering Criteria:} Gloss annotations for fingerspelling (e.g. `\textit{\textsc{\#a-n-n}}'). \\ \textbf{Manipulation:} Replace fingerspelled word with another contextually acceptable word. \end{tabular} & \begin{tabular}[t]{@{}p{3.5cm}@{}} \textbf{Matched}: \textcolor{greeen}{\textit{Ann}} hates fish but likes chicken. \\
\textbf{Mismatched}: \textcolor{reed}{\textit{Beth}} hates fish but likes chicken. \end{tabular} \\

\midrule
Classifiers ($N$=150) & Signs denoting the salient semantic properties of entities like size, shape, and number. Handshape tends to refer to entities and movement to events.  & \begin{tabular}[t]{@{}p{4.8cm}@{}} \textbf{Filtering Criteria:} Gloss annotations with "cl:" in their prefix.\\ \textbf{Manipulation:} Replace word signed as a classifier with another contextually acceptable word. \end{tabular} & \begin{tabular}[t]{@{}p{3.5cm}@{}} \textbf{Matched}: Are \textcolor{greeen}{\textit{the friends (with a two-handed DCL:C$^\dagger$)}} going out?  \\
\textbf{Mismatched}: Is \textcolor{reed}{\textit{the friend}} going out? \end{tabular} \\

\midrule

Conditional Statements ($N$=205) & 

\textit{If..., then...} statements. Conveyed with the manual sign \textsc{if} with eyebrow raise.
& \begin{tabular}[t]{@{}p{4.8cm}@{}} \textbf{Filtering Criteria:} Sentences containing the conditional marker "if". \\ \textbf{Manipulation:} Replace \textit{if} with \textit{when}. \end{tabular} & \begin{tabular}[t]{@{}p{3.5cm}@{}} \textbf{Matched}: \textcolor{greeen}{\textit{If}} I see my friend, I will be thrilled.  \\
\textbf{Mismatched}: \textcolor{reed}{\textit{When}} I see my friend, I will be thrilled. \end{tabular} \\

\midrule

Negation vs. Positive ($N$=104) & Expressions of negative polarity in a sentence, often through the manual sign \textsc{not} and the non-manual headshake. & \begin{tabular}[t]{@{}p{4.8cm}@{}} \textbf{Filtering Criteria:} Instances containing explicit negation markers (i.e. ``not'' or contractions with ``n't''). \\ \textbf{Manipulation:} Remove negation.\end{tabular} & \begin{tabular}[t]{@{}p{3.5cm}@{}} \textbf{Matched}: Bob \textcolor{greeen}{\textit{hasn't}} sent the letter. \\
\textbf{Mismatched}: Bob \textcolor{reed}{\textit{has}} sent the letter. \end{tabular} \\
\midrule

Positive vs. Negation ($N$=104) & Expressions of positive polarity in a sentence, without negative manual signs or headshake. & \begin{tabular}[t]{@{}p{4.8cm}@{}} \textbf{Filtering Criteria:} Instances not containing explicit negation markers (i.e. ``not'' or contractions with ``n't''). \\\textbf{Manipulation:} Add negation word.\end{tabular} & \begin{tabular}[t]{@{}p{3.5cm}@{}} \textbf{Matched}: Bob \textcolor{greeen}{\textit{read}} a book. \\
\textbf{Mismatched}: Bob \textcolor{reed}{\textit{didn't read}} a book. \end{tabular} \\
\midrule

Wh-Questions ($N$=123) & Content questions formed by wh- signs (what, who, where, when, why, how) and the non-manuals eyebrow lowering and/or head tilt. & \begin{tabular}[t]{@{}p{4.8cm}@{}} \textbf{Filtering Criteria:} Sentences ending with a ``?'' and containing at least one wh-word (``what'', ``when'', ``who'', ``where'', ``why'', ``whom'', or ``how''). \\ \textbf{Manipulation:} Replace wh-word with another acceptable wh-word. \end{tabular} & \begin{tabular}[t]{@{}p{3.5cm}@{}} \textbf{Matched}: \textcolor{greeen}{\textit{When}} did father arrive home? \\
\textbf{Mismatched}: \textcolor{reed}{\textit{How}} did father arrive home? \end{tabular} \\ 
\midrule
Polar Questions vs. Declaratives ($N$=150) & Polar questions formed only by the non-manual eyebrow raise. & \begin{tabular}[t]{@{}p{4.8cm}@{}} \textbf{Filtering Criteria:} Sentences ending with a ``?'', followed by manual verification for Polar Questions. \\ \textbf{Manipulation:} Convert to declarative. \end{tabular} & \begin{tabular}[t]{@{}p{3.5cm}@{}} \textbf{Matched}: \textcolor{greeen}{\textit{Are Jen and Joe married?}} \\ 
\textbf{Mismatched}: \textcolor{reed}{\textit{Jen and Joe are married.}} \end{tabular} \\
\midrule
Declaratives vs. Polar Questions ($N$=150) & Declarative sentences which do not involve any question. & \begin{tabular}[t]{@{}p{4.8cm}@{}} \textbf{Filtering Criteria:} Sentences that do not end with a ``?''. \\ \textbf{Manipulation:} Convert to polar question. \end{tabular} & \begin{tabular}[t]{@{}p{3.5cm}@{}} \textbf{Matched}: \textcolor{greeen}{\textit{All the men left together.}} \\ 
\textbf{Mismatched}: \textcolor{reed}{\textit{Did all the men leave together?}} \end{tabular} \\
\bottomrule
\end{tabular}}

\caption{Phenomena included in \dataset{}, along with their sample sizes, descriptions, construction, and examples. $^\dagger$ In the example for classifiers, we show the difference in classifiers using \asllrp{} notation: DCL - descriptive classifier, C - handshape. This classifier refers to a group of friends and cannot refer to a singular entity.}
\vspace{-1em}
\label{tab:phenomena}
\end{table*}

\paragraph{Dataset Construction}

Tab.~\ref{tab:phenomena} shows a detailed description of our stimuli design methods, across the 9 phenomena. 
Our general stimuli construction is as follows. First, we queried \asllrp{} for instances (consisting of a video, its glossed version, and its English representation) suited for a given phenomenon. Then, for each instance, we manipulated its English translation by replacing certain words or rewriting it to target the phenomenon in question. Taking ``Polar Questions vs. Declaratives'' as an example, we rewrote the matched sentence \textit{Are Jen and Joe married?} in its declarative form to create its mismatched counterpart: \textit{Jen and Joe are married.} Importantly, both the matched and mismatched utterances are grammatically correct---they differ in whether they are a correct translation of the input ASL video, and specifically in terms of the phenomenon in question. All of these considerations, applied to \asllrp{}, yield the focused dataset \dataset{} of 1,275 pairs across phenomena.

\subsection{Using \dataset{} for Sign-Conditioned Minimal Pair Analysis}
\label{sec:minimal_pair_analysis}

To analyze a model's behavior on the linguistic phenomena described above, we adopt standard practice in minimal-pair evaluation (see \cref{sec:eval-lit} for an overview), and compare the model's log-probabilities on the sentences in each pair, when conditioned on the ASL input.

Let $\mathcal{D} = \{(F_1, a_1, u_1), \dots, (F_n, a_n, u_n)\}$ be a phenomenon-specific dataset whose entries comprise input features extracted from the sign language video $F_i \in \mathbb{R}^{T \times d}$ (where $T$ is the number of frames) involving the phenomenon, a matched, ground-truth reference sentence translation of the video $a_i$, and a minimally differing sentence $u_i$ that has been perturbed in a targeted manner to be mismatched. We expect a model that has mastery over the target phenomenon to find the mismatched sentence $u_i$ more `surprising', or unlikely, than the matched sentence $a_i$, when conditioned on the video $F_i$. We measure the model's (un)likelihood for a sentence $s_i := (x_1, \dots, x_{|s_i|})$ by computing its conditional, per-token surprisal (negative log-probability):
\begin{align}
    \mathcal{S}(s_i) = \frac{1}{\left|s_i\right|}\sum_{t=1}^{\left|s_i\right|}-\log p(x_t \mid x_{<t}, F_i)
\end{align}
We then compute the difference in surprisals for the mismatched and matched sentences:
\begin{align}
\label{eq:surpdiff}
    \Delta{}\text{Surprisal}_i = \mathcal{S}(u_i) - \mathcal{S}(a_i)
\end{align}

Insofar as a model is sensitive to the phenomenon that governs the differences between $a_i$ and $u_i$, we expect $\Delta{}\text{Surprisal}_i$ to be greater than 0. Accuracy, then, is the proportion of pairs for which $\Delta{}\text{Surprisal} > 0$. Since this comparison is done over pairs, chance performance is 50\%.

The basic use case we envision for \datasetshort{}, then, is to evaluate accuracy of an ASL-to-English translation model (in the sense of accuracy defined above) on the 9 phenomena-specific subsets.  This framework provides a general method to evaluate a model on a number of phenomena for which minimal pairs can be created, given a fixed video.  In our own case study (\cref{sec:case_study}), we further divide 2 of the 9 phenomena into two subsets each, corresponding to those examples that rely on non-manuals only vs.~both manuals and non-manuals.

\begin{figure*}[t!]
    \centering
    \resizebox{0.8\linewidth}{!}{%
        \includegraphics{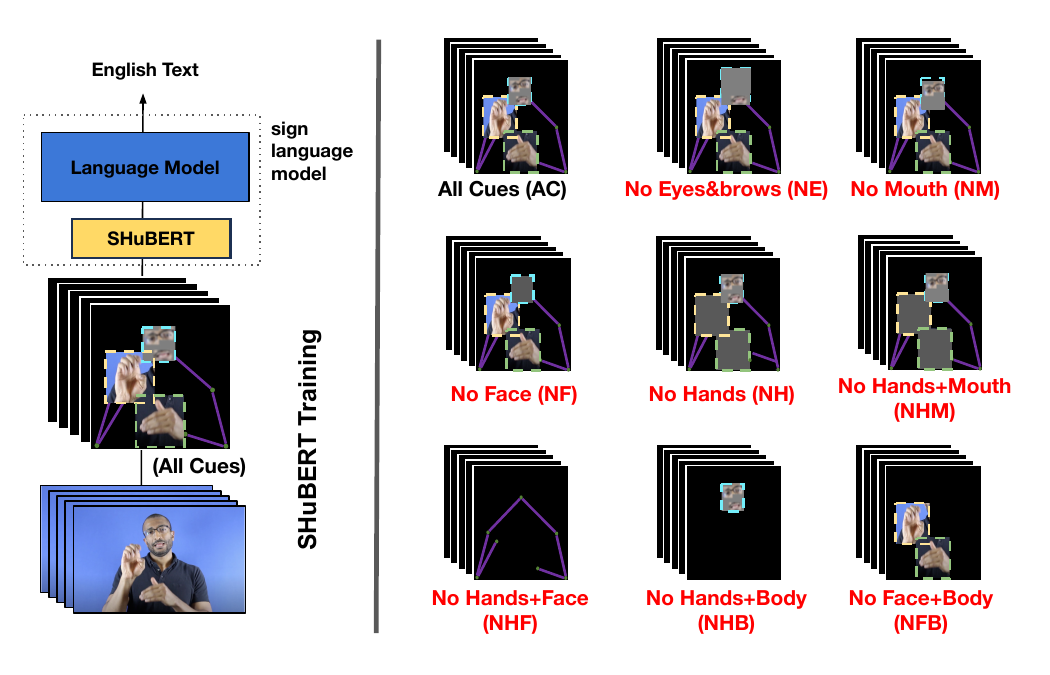}
    }
    \vspace{-20pt}
    \caption{\textbf{Left:} A depiction of how \shubert{} \citep{gueuwou2024shubertselfsupervisedsignlanguage} is combined with an off-the-shelf language model (here, ByT5) to perform ASL-to-English translation. \textbf{Right:} Examples of inputs provided to the model for the All Cues condition as well as the 8 Cue Ablations.}
    \label{fig:cue_ablation}
\end{figure*}

\section{A Case Study}
\label{sec:case_study}

Next we use \dataset{} for a case study, in which we analyze an open, state-of-the-art ASL-to-English translation model.  Specifically, we use \dataset{} to (1) study the model's behavior on the 9 phenomena, in terms of the surprisal-based accuracies defined above, and (2) analyze the extent to which the model uses cues from the hand, body, and facial information channels.

\subsection{Model studied}
\label{sec:model_studied}

Based on our research goal above, we define a set of criteria that govern our choice of model: First, the model must take in ASL video input and produce English translations, in a manner that allows the extraction of token probabilities (to facilitate analysis via surprisals). Second, it should enable control over the input channels that encode information from the cues we aim to study---e.g., one should be able to mask out information from the eyes while preserving other cues (hands, mouth, body movements).\footnote{We note that this criterion is not necessary to use \datasetshort{} or to do surprisal-based analysis, but only to carry out our case study which concerns the analysis of cue use.} Finally, the model weights, training data, and training pipeline should be openly available, and runnable on academic compute.

The only currently available model that meets these criteria is the \shubert{}+ByT5 translation model from \citet{gueuwou2024shubertselfsupervisedsignlanguage}.\footnote{\url{https://shubert.pals.ttic.edu/}}  This translation model combines two jointly fine-tuned models:  \shubert{}, a BERT-style \citep{devlin-etal-2019-bert} encoder pretrained on 1,000 hours of continuous ASL YouTube videos (combining subsets of YouTube-ASL~\cite{uthus2023youtube} and YouTube-SL-25~\cite{tanzer2024youtubesl25}), and the ByT5-Base text translation model~\cite{xue-etal-2022-byt5}. 
 
\shubert{} takes inputs that are decomposed into four channels: face (mouth and eye image crops), left hand crops, right hand crops, and body pose keypoints.  The face and hand crops are represented using DINOv2 image features~\cite{oquab2024dinov}. The input channels can be manipulated in order to measure \shubert{}'s use of information restricted to a particular channel. The translation model is trained to map from ASL videos to English translations, using the next-token prediction objective (here, the tokens are bytes) in an autoregressive manner. This means that we can use its next-token probabilities to compute the log-probabilities needed for our minimal pair analysis (see \cref{sec:minimal_pair_analysis})
In all of our experiments, we either (1) use the translation model as is while manipulating the input cues (i.e., at inference time) or (2) re-train versions of \shubert{} with varying cues present, which we again jointly fine-tune with ByT5, directly following the training pipeline (including hyperparameters) of~\citet{gueuwou2024shubertselfsupervisedsignlanguage}. 

\subsection{Cue Ablation}
\label{sec:cue_ablation_strategy}

To understand the importance of different visual cues to the model's performance, we use a systematic ablation strategy that masks specific features (corresponding to targeted cues) in the input video frames. For example, if masking the hands does not affect performance on a particular phenomenon, this indicates that the model is not sensitive to handshape and orientation (i.e., does not use these cues in predicting the token probabilities) in examples of that phenomenon.\footnote{In this case, the model may still be sensitive to hand {\it location}, which is encoded in the body pose.} We use the keypoints returned from the MediaPipe library \cite{lugaresi2019mediapipe} to detect the regions of interest, which are then selectively greyed out in the video frames (see~\cref{fig:cue_ablation}). We perform these ablations either at inference time (\cref{sec:inference-ablation}) or during training (\cref{sec:training-ablation}).

We use the original \shubert{}+ByT5 model with the full video input as our baseline condition, where no masking is performed, and refer to this as \textbf{All Cues (AC)}. Then, to isolate the impact of specific cues, we consider 8 ablations: 
1) \textbf{No Eyes \& Brows (NE)}, where the eye and eyebrow regions (involved in questions \citep{baker1983microanalysis} and conditionals \citep{liddell1980american, wilbur1999syntactic}) are masked from the face channel; 
2) \textbf{No Mouth (NM)}, where the mouth region is masked from the face channel, thereby removing mouthing cues, which are often used for disambiguation, or convey adjectival or adverbial meanings; 
3) \textbf{No Face (NF)}, where the entire face channel (eyes, eyebrows, and mouth) is masked;
4) \textbf{No Hands (NH)}, where the hand channels are masked;
5) \textbf{No Hands \& Mouth (NHM)}, where the mouth region of the face channel and the hand channels are masked; 
6) \textbf{No Hands \& Face (NHF)}, where the hand and face channels are masked, thereby allowing us to test if the model can use body pose alone, which is used to indicate role shifting, contrast, and spatial organization;
7) \textbf{No Hands \& Body (NHB)}, where only the face channel is retained; and 8) \textbf{No Face \& Body (NFB)}, where only the hand channels are retained.

\subsection{Experiment 1: Effect of Cue Ablations on Minimal Translation Pair Performance}
\label{sec:inference-ablation}

\begin{figure*}[!t]
    \centering
\includegraphics[width=\textwidth]{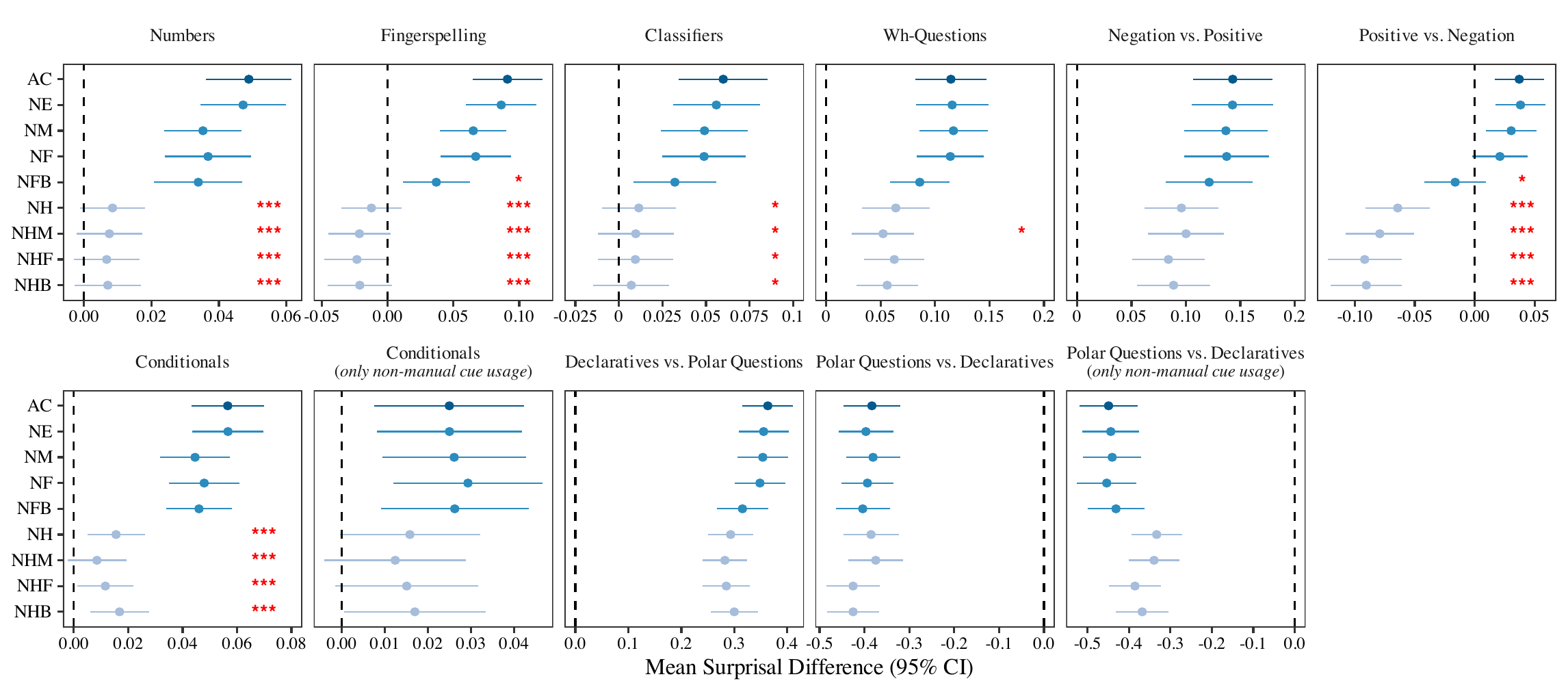}
    \caption{Average difference in surprisal of mismatched and matched sentences across phenomena and across \textbf{inference cue ablations}. Error bars indicate 95\% confidence intervals. Stars (*) indicate significance test results for comparing the surprisal difference in a given cue ablation to surprisal difference in the `All Cues' condition (AC). *: \textit{p <} .05; **: \textit{p <} .01; ***: \textit{p <} .001}
    \label{fig:inf-diff}
\end{figure*}

Our first experiment evaluates \shubert{}+ByT5 on \dataset{}, specifically focusing on the effect of ablating input cues at inference time, as discussed in \cref{sec:cue_ablation_strategy}. 
We quantify the extent to which a model is sensitive to a given set of cues by comparing its performance when those cues are ablated to performance in the All Cues (AC) condition. 
Tab.~\ref{tab:acc-inference} shows the phenomenon-specific accuracies obtained by \shubert{}+ByT5, each corresponding to a particular cue ablation. We split the `Conditionals' and `Polar Questions vs.~Declaratives' subsets of \dataset{} into two rows each---one corresponding to videos where the phenomenon is represented \textit{only} using non-manual cues, and the other where both manual and non-manual cues are used, giving us a total of 11 subsets.  We do this to enable finer-grained analysis of model performance on cases exclusively requiring sensitivity to non-manuals. Fig.~\ref{fig:inf-diff} shows average $\Delta\text{Surprisal}$ values across all phenomena and channel ablations.

\paragraph{Results with all cues} We first summarize the results in the ``All Cues'' condition, as this serves as the baseline against which we will compare subsequent cue ablation results. We find that the model performs above chance (50\%) on 9 out of 11 subsets, showing particularly good performance on Numbers, Fingerspelling, Wh-Questions, and Negation vs.~Positive, while performing substantially below chance on both ``Polar Questions vs.~Declaratives'' subsets (we return to this below). There is no clear relationship between performance and the primary or secondary cue(s) involved in a phenomenon, nor is there one between performance and the cardinality of the categories involved; for example, Numbers and Fingerspelling have large vocabularies while Negation vs.~Positive is a binary distinction, and all of these have among the highest performance.  Among the phenomena on which the model performs above chance, it performs the worst on Classifiers. 
Classifier meanings critically rely on referents within the utterance~\cite{zwitserlood2012classifiers, hakguder2021}, suggesting that analyses of classifier sign errors and reference may be a good direction for future work.

\begin{table*}[!t]
    \centering
    \resizebox{\textwidth}{!}{
    \begin{tabular}{@{}lllrccccccccc@{}}
\toprule
\textbf{Phenomenon} & \textbf{Primary Cue(s)} & \textbf{Secondary Cue} & \textbf{\#} & \textbf{AC} & \textbf{NE} & \textbf{NM} & \textbf{NF} & \textbf{NFB} & \textbf{NH} & \textbf{NHM} & \textbf{NHF} & \textbf{NHB}\\
\midrule
1. Numbers & Hands &  & 119 & \cellcolor[HTML]{1b7378}{0.87} & \cellcolor[HTML]{548f8a}{0.82} & \cellcolor[HTML]{548f8a}{0.78} & \cellcolor[HTML]{548f8a}{\textbf{0.77}} & \cellcolor[HTML]{548f8a}{\textbf{0.75}} & \cellcolor[HTML]{80ac9c}{\textbf{0.61}} & \cellcolor[HTML]{95baa5}{\textbf{0.58}} & \cellcolor[HTML]{95baa5}{\textbf{0.55}} & \cellcolor[HTML]{95baa5}{\textbf{0.55}}\\
2. Fingerspelling & Hands &  & 170 & \cellcolor[HTML]{548f8a}{0.78} & \cellcolor[HTML]{548f8a}{0.75} & \cellcolor[HTML]{80ac9c}{\textbf{0.70}} & \cellcolor[HTML]{80ac9c}{0.72} & \cellcolor[HTML]{80ac9c}{\textbf{0.68}} & \cellcolor[HTML]{95baa5}{\textbf{0.49}} & \cellcolor[HTML]{95baa5}{\textbf{0.46}} & \cellcolor[HTML]{aac9af}{\textbf{0.43}} & \cellcolor[HTML]{95baa5}{\textbf{0.49}}\\
3. Classifiers & Hands &  & 150 & \cellcolor[HTML]{80ac9c}{0.63} & \cellcolor[HTML]{80ac9c}{0.62} & \cellcolor[HTML]{80ac9c}{0.59} & \cellcolor[HTML]{80ac9c}{0.59} & \cellcolor[HTML]{95baa5}{\textbf{0.53}} & \cellcolor[HTML]{95baa5}{\textbf{0.51}} & \cellcolor[HTML]{95baa5}{\textbf{0.48}} & \cellcolor[HTML]{95baa5}{\textbf{0.49}} & \cellcolor[HTML]{95baa5}{\textbf{0.47}}\\
4. Wh-Questions & Hands + Brow lowered & Head shake & 123 & \cellcolor[HTML]{548f8a}{0.75} & \cellcolor[HTML]{548f8a}{0.74} & \cellcolor[HTML]{548f8a}{0.76} & \cellcolor[HTML]{548f8a}{0.76} & \cellcolor[HTML]{80ac9c}{0.72} & \cellcolor[HTML]{80ac9c}{\textbf{0.66}} & \cellcolor[HTML]{80ac9c}{\textbf{0.65}} & \cellcolor[HTML]{80ac9c}{0.67} & \cellcolor[HTML]{80ac9c}{\textbf{0.64}}\\
5. Negation vs. Positive & Hands + Head shake &  & 104 & \cellcolor[HTML]{548f8a}{0.80} & \cellcolor[HTML]{548f8a}{0.77} & \cellcolor[HTML]{548f8a}{0.77} & \cellcolor[HTML]{548f8a}{0.77} & \cellcolor[HTML]{80ac9c}{0.71} & \cellcolor[HTML]{80ac9c}{0.69} & \cellcolor[HTML]{80ac9c}{0.69} & \cellcolor[HTML]{80ac9c}{0.68} & \cellcolor[HTML]{80ac9c}{0.69}\\
6. Positive vs. Negation & Hands &  & 104 & \cellcolor[HTML]{80ac9c}{0.65} & \cellcolor[HTML]{80ac9c}{0.68} & \cellcolor[HTML]{80ac9c}{0.62} & \cellcolor[HTML]{80ac9c}{0.66} & \cellcolor[HTML]{95baa5}{\textbf{0.53}} & \cellcolor[HTML]{aac9af}{\textbf{0.34}} & \cellcolor[HTML]{aac9af}{\textbf{0.34}} & \cellcolor[HTML]{bfd8b8}{\textbf{0.28}} & \cellcolor[HTML]{bfd8b8}{\textbf{0.29}}\\
7a. Conditionals & Hands + Brow raise & Head thrust, Body forward & 155 & \cellcolor[HTML]{80ac9c}{0.70} & \cellcolor[HTML]{80ac9c}{0.72} & \cellcolor[HTML]{80ac9c}{0.66} & \cellcolor[HTML]{80ac9c}{0.70} & \cellcolor[HTML]{80ac9c}{0.72} & \cellcolor[HTML]{95baa5}{\textbf{0.56}} & \cellcolor[HTML]{95baa5}{\textbf{0.51}} & \cellcolor[HTML]{95baa5}{\textbf{0.54}} & \cellcolor[HTML]{80ac9c}{\textbf{0.59}}\\
7b. Conditionals (\textit{NM only}) & Brow raise & Head thrust, Body forward & 50 & \cellcolor[HTML]{80ac9c}{0.68} & \cellcolor[HTML]{80ac9c}{0.68} & \cellcolor[HTML]{80ac9c}{0.68} & \cellcolor[HTML]{80ac9c}{0.70} & \cellcolor[HTML]{548f8a}{0.76} & \cellcolor[HTML]{80ac9c}{0.60} & \cellcolor[HTML]{80ac9c}{0.62} & \cellcolor[HTML]{80ac9c}{0.60} & \cellcolor[HTML]{80ac9c}{0.64}\\
8. Declaratives vs.~Polar Questions & Hands &  & 150 & \cellcolor[HTML]{1b7378}{0.97} & \cellcolor[HTML]{1b7378}{0.97} & \cellcolor[HTML]{1b7378}{0.97} & \cellcolor[HTML]{1b7378}{0.96} & \cellcolor[HTML]{1b7378}{0.96} & \cellcolor[HTML]{1b7378}{0.96} & \cellcolor[HTML]{1b7378}{0.95} & \cellcolor[HTML]{1b7378}{0.95} & \cellcolor[HTML]{1b7378}{0.95}\\
9a. Polar Qs vs. Declaratives & Hands + Brow raise & Head forward & 57 & \cellcolor[HTML]{e9f7cb}{0.04} & \cellcolor[HTML]{e9f7cb}{0.04} & \cellcolor[HTML]{e9f7cb}{0.04} & \cellcolor[HTML]{e9f7cb}{0.05} & \cellcolor[HTML]{e9f7cb}{0.04} & \cellcolor[HTML]{e9f7cb}{0.05} & \cellcolor[HTML]{e9f7cb}{0.05} & \cellcolor[HTML]{e9f7cb}{0.04} & \cellcolor[HTML]{e9f7cb}{0.04}\\
9b. Polar Qs vs. Declaratives (\textit{NM only}) & Brow raise & Head forward & 93 & \cellcolor[HTML]{e9f7cb}{0.09} & \cellcolor[HTML]{e9f7cb}{0.09} & \cellcolor[HTML]{e9f7cb}{0.09} & \cellcolor[HTML]{e9f7cb}{0.09} & \cellcolor[HTML]{e9f7cb}{0.10} & \cellcolor[HTML]{e9f7cb}{0.15} & \cellcolor[HTML]{e9f7cb}{0.14} & \cellcolor[HTML]{e9f7cb}{0.09} & \cellcolor[HTML]{e9f7cb}{0.13}\\
\bottomrule
\end{tabular}
    }
    \caption{Phenomenon-wise surprisal-derived accuracies across \textbf{inference conditions}. Accuracy values are \textbf{boldfaced} if they are significantly different from the accuracy on the `All Cues' (AC) inference condition ($p <$ .05, as measured by a two-tailed exact binomial test, with the Bonferroni correction for multiple comparisons). ``(\textit{NM only})'' indicates that the stimuli in that subset involve only non-manual cues.  ``Hands'' refers to any number of cues related to handshape and orientation.  Chance performance is 50\%.}
    \label{tab:acc-inference}
\end{table*}

\paragraph{Heavy reliance on hands} 
When the hands are ablated (i.e., in the NH, NHM, NHF, and NHB conditions), the model performs significantly worse than in the AC condition (often worse than or close to chance) on Numbers, Fingerspelling, Classifiers, Wh-Questions, Positive vs.~Negation, and Conditionals (NM only). This result is expected, as hands are the most important source of information in sign language in general \cite{malaia2018information} as well as a primary cue for these phenomena.\footnote{When we remove hands but not body pose (NH, NHM, NHF), we retain information about hand location from the body pose, but lose important handshape/orientation features.}  When hands are not a primary cue (e.g.~for the Conditionals (NM only) subset), there is no significant performance reduction when ablating the hands.

\begin{table*}[!t]
    \centering
    \resizebox{\textwidth}{!}{
    \begin{tabular}{lllrccccccccc}
\toprule
\textbf{Phenomenon} & \textbf{Primary Cue(s)} & \textbf{Secondary Cue} & \textbf{\#} & \textbf{AC} & \textbf{NE} & \textbf{NM} & \textbf{NF} & \textbf{NFB} & \textbf{NH} & \textbf{NHM} & \textbf{NHF} & \textbf{NHB}\\
\midrule
1. Numbers & Hands &  & 119 & \cellcolor[HTML]{95baa5}{0.50} & \cellcolor[HTML]{95baa5}{0.50} & \cellcolor[HTML]{95baa5}{0.47} & \cellcolor[HTML]{95baa5}{0.48} & \cellcolor[HTML]{aac9af}{\textbf{0.43}} & \cellcolor[HTML]{bfd8b8}{\textbf{0.28}} & \cellcolor[HTML]{bfd8b8}{\textbf{0.28}} & \cellcolor[HTML]{bfd8b8}{\textbf{0.26}} & \cellcolor[HTML]{aac9af}{\textbf{0.32}}\\
2. Fingerspelling & Hands &  & 170 & \cellcolor[HTML]{aac9af}{0.43} & \cellcolor[HTML]{aac9af}{0.42} & \cellcolor[HTML]{aac9af}{\textbf{0.40}} & \cellcolor[HTML]{aac9af}{0.41} & \cellcolor[HTML]{aac9af}{\textbf{0.35}} & \cellcolor[HTML]{bfd8b8}{\textbf{0.29}} & \cellcolor[HTML]{bfd8b8}{\textbf{0.26}} & \cellcolor[HTML]{bfd8b8}{\textbf{0.26}} & \cellcolor[HTML]{bfd8b8}{\textbf{0.29}}\\
3. Classifiers & Hands &  & 150 & \cellcolor[HTML]{95baa5}{0.44} & \cellcolor[HTML]{aac9af}{0.43} & \cellcolor[HTML]{aac9af}{\textbf{0.41}} & \cellcolor[HTML]{aac9af}{\textbf{0.41}} & \cellcolor[HTML]{aac9af}{\textbf{0.36}} & \cellcolor[HTML]{bfd8b8}{\textbf{0.28}} & \cellcolor[HTML]{bfd8b8}{\textbf{0.26}} & \cellcolor[HTML]{bfd8b8}{\textbf{0.26}} & \cellcolor[HTML]{bfd8b8}{\textbf{0.29}}\\
4. Wh-Questions & Hands + Brow lower & Head shake & 123 & \cellcolor[HTML]{95baa5}{0.54} & \cellcolor[HTML]{95baa5}{0.53} & \cellcolor[HTML]{95baa5}{0.53} & \cellcolor[HTML]{95baa5}{0.55} & \cellcolor[HTML]{95baa5}{\textbf{0.44}} & \cellcolor[HTML]{bfd8b8}{\textbf{0.30}} & \cellcolor[HTML]{bfd8b8}{\textbf{0.23}} & \cellcolor[HTML]{bfd8b8}{\textbf{0.26}} & \cellcolor[HTML]{bfd8b8}{\textbf{0.28}}\\
5. Negation vs.~Positive & Hands + Head shake &  & 104 & \cellcolor[HTML]{95baa5}{0.51} & \cellcolor[HTML]{95baa5}{0.50} & \cellcolor[HTML]{95baa5}{\textbf{0.45}} & \cellcolor[HTML]{95baa5}{0.47} & \cellcolor[HTML]{aac9af}{\textbf{0.40}} & \cellcolor[HTML]{bfd8b8}{\textbf{0.26}} & \cellcolor[HTML]{bfd8b8}{\textbf{0.25}} & \cellcolor[HTML]{bfd8b8}{\textbf{0.25}} & \cellcolor[HTML]{bfd8b8}{\textbf{0.29}}\\
6. Positive vs.~Negation & Hands &  & 104 & \cellcolor[HTML]{95baa5}{0.47} & \cellcolor[HTML]{95baa5}{0.47} & \cellcolor[HTML]{95baa5}{0.44} & \cellcolor[HTML]{95baa5}{0.45} & \cellcolor[HTML]{aac9af}{\textbf{0.39}} & \cellcolor[HTML]{bfd8b8}{\textbf{0.29}} & \cellcolor[HTML]{bfd8b8}{\textbf{0.26}} & \cellcolor[HTML]{bfd8b8}{\textbf{0.24}} & \cellcolor[HTML]{bfd8b8}{\textbf{0.29}}\\
7a. Conditionals & Hands + Brow raise & Head thrust, Body forward & 155 & \cellcolor[HTML]{95baa5}{0.45} & \cellcolor[HTML]{95baa5}{0.45} & \cellcolor[HTML]{aac9af}{0.43} & \cellcolor[HTML]{aac9af}{0.43} & \cellcolor[HTML]{aac9af}{\textbf{0.38}} & \cellcolor[HTML]{bfd8b8}{\textbf{0.28}} & \cellcolor[HTML]{bfd8b8}{\textbf{0.25}} & \cellcolor[HTML]{bfd8b8}{\textbf{0.27}} & \cellcolor[HTML]{aac9af}{\textbf{0.31}}\\
7b. Conditionals (\textit{NM only}) & Brow raise & Head thrust, Body forward & 50 & \cellcolor[HTML]{aac9af}{0.41} & \cellcolor[HTML]{aac9af}{0.40} & \cellcolor[HTML]{aac9af}{0.41} & \cellcolor[HTML]{aac9af}{0.39} & \cellcolor[HTML]{aac9af}{0.37} & \cellcolor[HTML]{bfd8b8}{\textbf{0.26}} & \cellcolor[HTML]{bfd8b8}{\textbf{0.25}} & \cellcolor[HTML]{bfd8b8}{\textbf{0.26}} & \cellcolor[HTML]{bfd8b8}{\textbf{0.27}}\\
8. Declaratives vs.~Polar Qs & Hands &  & 150 & \cellcolor[HTML]{95baa5}{0.51} & \cellcolor[HTML]{95baa5}{0.50} & \cellcolor[HTML]{95baa5}{0.48} & \cellcolor[HTML]{95baa5}{0.48} & \cellcolor[HTML]{aac9af}{\textbf{0.40}} & \cellcolor[HTML]{bfd8b8}{\textbf{0.27}} & \cellcolor[HTML]{bfd8b8}{\textbf{0.24}} & \cellcolor[HTML]{bfd8b8}{\textbf{0.24}} & \cellcolor[HTML]{bfd8b8}{\textbf{0.29}}\\
9a. Polar Qs vs.~Declaratives & Hands + Brow raise & Head forward & 57 & \cellcolor[HTML]{aac9af}{0.41} & \cellcolor[HTML]{aac9af}{0.41} & \cellcolor[HTML]{aac9af}{0.39} & \cellcolor[HTML]{aac9af}{0.40} & \cellcolor[HTML]{aac9af}{0.36} & \cellcolor[HTML]{bfd8b8}{\textbf{0.21}} & \cellcolor[HTML]{bfd8b8}{\textbf{0.21}} & \cellcolor[HTML]{bfd8b8}{\textbf{0.23}} & \cellcolor[HTML]{bfd8b8}{\textbf{0.23}}\\
9b. Polar Qs vs.~Declaratives (\textit{NM only}) & Brow raise & Head forward & 93 & \cellcolor[HTML]{95baa5}{0.52} & \cellcolor[HTML]{95baa5}{0.50} & \cellcolor[HTML]{95baa5}{0.48} & \cellcolor[HTML]{95baa5}{0.49} & \cellcolor[HTML]{95baa5}{\textbf{0.45}} & \cellcolor[HTML]{bfd8b8}{\textbf{0.20}} & \cellcolor[HTML]{bfd8b8}{\textbf{0.17}} & \cellcolor[HTML]{bfd8b8}{\textbf{0.25}} & \cellcolor[HTML]{bfd8b8}{\textbf{0.25}}\\
\bottomrule
\end{tabular}}
    \caption{Phenomenon-wise BLEURT scores across \textbf{inference conditions}. Values are \textbf{boldfaced} if they are significantly different from the BLEURT score in the `All Cues' (AC) inference condition ($p <$ .05, as measured by a $t$-test, with the Bonferroni correction for multiple comparisons). 
    ``(\textit{NM only})'' indicates that the stimuli in that subset involve only non-manual cues.  ``Hands'' refers to any number of cues related to handshape and orientation.}
    \label{tab:bleurt-cues}
    \vspace{-.6em}
\end{table*}

\paragraph{Poor sensitivity to non-manual cues} The model is much less sensitive to non-manual cues (e.g., head movements, eyebrow raises), even on phenomena that explicitly rely on these cues---Wh-Questions, Negation vs.~Positive, Conditionals, and Polar Questions vs.~Declaratives. That is, its accuracies on these phenomena are no different from those in the AC condition when these cues are ablated. For example, on the subset of the Conditionals that \textit{exclusively} rely on non-manual cues, we notice model insensitivity across \textit{all} cue-ablation conditions, but this could also be due to the relatively small sample size (50). The only cases where we do observe sensitivity to non-manual cues are in phenomena that do not necessarily rely on them---e.g., the model is significantly worse (relative to AC) in the absence of the face and body pose (NF and/or NFB) for Numbers, Fingerspelling, Classifiers, and Positive vs.~Negation.  This could be because mouthing is sometimes used to disambiguate certain signs, even when this is not a necessary cue, and the presence of mouthing is not annotated in the data. 
Overall, the model is not as sensitive to critical non-manual cues as we might expect from linguistic intuition, despite having access to these cues during training.

\paragraph{Declarative bias} 
Among the phenomena tested, there was a particularly wide gap between model accuracies on ``Declaratives vs. Polar Questions'' and ``Polar Questions vs. Declaratives'' in {\it all} experimental conditions. In particular, the model shows a bias towards generating declarative sentences over polar questions, in all cue ablation settings (seen in both~\Cref{tab:acc-inference} and the $\Delta\text{Surprisal}$ results in Fig.~\ref{fig:inf-diff}.

\subsection{Experiment 2: Surprisals vs.~BLEURT}
\label{sec:bleurt}

The results thus far suggest that \shubert{}+ByT5 is not always sensitive to the various cues it is trained to use. While we base these findings on our proposed surprisal analysis, to what extent could they also be explained using standard machine translation (MT) metrics, like \bleurt{} \citep{sellam2020bleurt}? That is, what does the minimal pair analysis buy us above and beyond off-the-shelf translation measures?

An a priori argument against \bleurt{} (and other general MT metrics) is that it is a global similarity measure between reference and translation, and is not guaranteed to be systematically sensitive towards the phenomena in \dataset{}. Taking Wh-Questions as an example, a model might succeed at recognizing the right Wh-word but produce a completely wrong translation: In one example (in the All Cues condition), where the reference is \textit{\textbf{Why} do you have to move out of San Diego?}, the model produces \textit{\textbf{Why} do you think this is ASL?}. Here, the BLEURT score is poor (24.4) but it is not due to an insensitivity to the Wh-word, but instead due to the completely different meanings encoded in the two sentences because of other word substitutions.

To confirm our a priori intuition, we obtain hypothesized translations of the \dataset{} inputs from the \shubert{}+ByT5 model (using beam search, as in~\citet{gueuwou2024shubertselfsupervisedsignlanguage}), and compute the average \bleurt{} scores between the model translations and the ground-truth reference sentences. We report the resulting BLEURT scores across phenomena and cue ablations in Tab.~\ref{tab:bleurt-cues}.  

These results indeed confirm that BLEURT cannot uncover the distinctions in model behavior across phenomena and input conditions that we found in the minimal pair analysis.  First, there is very little variability in BLEURT scores across phenomena.  Presumably, as in our example above, BLEURT is dominated by various differences in translations besides the specific ones we target.  Second, for most phenomena BLEURT is lower whenever the hands or body are removed, again with little distinction among phenomena.  BLEURT is therefore unable to discover the model's insensitivity to certain non-manual cues in some phenomena.  Finally, we also measure the Pearson correlation between BLEURT and surprisal-based accuracy for each phenomenon, and find generally poor to moderate correlations (ranging from -.17 for 'Polar Questions vs.~Declaratives' to .36 for Numbers).  
These results are not surprising, but reinforce the role of minimal pair analysis, which can help diagnose specific, linguistically interpretable model behaviors that translation metrics do not reveal.

\begin{table*}[]
\centering
\resizebox{0.8\textwidth}{!}{
\begin{tabular}{lllrlll}
\toprule
\textbf{Phenomenon} & \textbf{Primary Cue(s)} & \textbf{Secondary Cue} & \textbf{\#} & \textbf{AC} & \textbf{NF} & \textbf{NFB}\\
\midrule
1. Numbers & Hands &  & 119 & \cellcolor[HTML]{1b7378}{0.87} & \cellcolor[HTML]{548f8a}{\textbf{0.76}} & \cellcolor[HTML]{548f8a}{\textbf{0.73}}\\
2. Fingerspelling  & Hands &  & 170 & \cellcolor[HTML]{548f8a}{0.78} & \cellcolor[HTML]{548f8a}{0.74} & \cellcolor[HTML]{80ac9c}{\textbf{0.64}}\\
3. Classifiers & Hands &  & 150 & \cellcolor[HTML]{80ac9c}{0.63} & \cellcolor[HTML]{80ac9c}{0.63} & \cellcolor[HTML]{95baa5}{0.58}\\
4. Wh-Questions  & Hands + Brow lower & Head shake & 123 & \cellcolor[HTML]{548f8a}{0.75} & \cellcolor[HTML]{95baa5}{\textbf{0.54}} & \cellcolor[HTML]{80ac9c}{\textbf{0.63}}\\
5. Negation vs.~Positive  & Hands + Head shake &  & 104 & \cellcolor[HTML]{548f8a}{0.80} & \cellcolor[HTML]{548f8a}{0.81} & \cellcolor[HTML]{548f8a}{0.76}\\
6. Positive vs.~Negation  & Hands &  & 104 & \cellcolor[HTML]{80ac9c}{0.65} & \cellcolor[HTML]{80ac9c}{0.62} & \cellcolor[HTML]{80ac9c}{0.62}\\
7a. Conditionals & Hands + Brow raise & Head thrust, Body forward & 155 & \cellcolor[HTML]{80ac9c}{0.70} & \cellcolor[HTML]{1b7378}{\textbf{0.89}} & \cellcolor[HTML]{80ac9c}{0.61}\\
7b. Conditionals \textit{(NM Only)} & Brow raise & Head thrust, Body forward & 50 & \cellcolor[HTML]{80ac9c}{0.68} & \cellcolor[HTML]{80ac9c}{0.72} & \cellcolor[HTML]{95baa5}{0.46}\\
8. Declaratives vs.~Polar Questions  & Hands &  & 150 & \cellcolor[HTML]{1b7378}{0.97} & \cellcolor[HTML]{1b7378}{0.96} & \cellcolor[HTML]{1b7378}{0.95}\\
9a. Polar Qs vs.~Declaratives & Hands + Brow raise &  & 57 & \cellcolor[HTML]{e9f7cb}{0.04} & \cellcolor[HTML]{e9f7cb}{0.09} & \cellcolor[HTML]{e9f7cb}{0.07}\\
9b. Polar Qs vs.~Declaratives \textit{(NM Only)} & Brow raise & Head forward & 93 & \cellcolor[HTML]{e9f7cb}{0.09} & \cellcolor[HTML]{e9f7cb}{0.11} & \cellcolor[HTML]{e9f7cb}{0.04}\\
\bottomrule
\end{tabular}
}
    \caption{Phenomenon-wise accuracies across \textbf{training conditions}.  The input channels match the training condition in each column.  Values are \textbf{boldfaced} if they are significantly different from the accuracy in the `All Cues' (AC) inference condition ($p <$ .05, as measured by a two-tailed exact binomial test, with the Bonferroni correction for multiple comparisons). ``(\textit{NM only})'' indicates that the stimuli in that subset only involved the usage of non-manual cues.  ``Hands'' refers to any number of cues related to handshape and orientation. Chance performance is 50\%.}
    \label{tab:acc-training}
\end{table*}

\subsection{Experiment 3: Controlled Rearing of \shubert{} Using Cue Ablations}
\label{sec:training-ablation}

There are multiple possible explanations for the results of the inference-time cue ablation analysis of Experiment 1.  One possibility is that the ablated inputs are out of distribution for the model, since it is trained on the full set of cues, so we may not know how the model would behave if it were both trained and tested with the ablated input.  In particular, one may wonder whether decreased performance when cues are ablated is due to the train-test mismatch and not due to the missing cue information.

To address this issue, we run ``controlled rearing'' experiments \citep[][a.o.]{jumelet-etal-2021-language, misra-mahowald-2024-language, leong-linzen-2023-language}, where we train new variants of \shubert{} with certain channels removed during training:  We mask out the features of the ablated channel(s), re-train \shubert{} using the masked input, then combine it with ByT5 and fine-tune as for the original \shubert{}+ByT5.  We follow the fine-tuning recipe in~\citet{gueuwou2024shubertselfsupervisedsignlanguage}:  We first fine-tune on a large corpus of weakly aligned ASL-English pairs ($\sim$800K samples from the union of YouTube-ASL \cite{uthus2023youtube} and the ASL part of YouTube-SL-25 \cite{tanzer2024youtubesl25}), and then continue fine-tuning on a smaller ($\sim$200K samples) but more accurately aligned training set  consisting of How2Sign \cite{duarte2021how2sign}, ASL Stem Wiki \cite{yin2024asl}, and OpenASL \cite{shi2022open}.

We conduct this experiment for two types of channel ablations: One where the face is removed (NF), and one where only the hands are retained (NFB). We perform our surprisal-based minimal pair analysis, and compare accuracies in these conditions to those in the AC condition. Tab.~\ref{tab:acc-training} shows the phenomenon-specific accuracies, while Fig.~\ref{fig:training-diff} in \cref{sec:complementary} shows average $\Delta\text{Surprisal}$ values.

For most phenomena, we see similar relative changes from the AC condition, and the models are still susceptible to declarative bias.  While there are some differences (notably for Wh-Questions and Conditionals), there is no consistent improvement in performance between the inference-time ablations and the controlled rearing setting.  This suggests that our inference-time ablation results are not explained away by train-test mismatch.  Additional investigation of the reasons behind differences across phenomena is left for future work.

\section{Conclusion}
\label{sec:conclusion}

\dataset{} is, to our knowledge, the first dataset for minimal pair analysis of linguistic phenomena in sign language translation models. We designed \dataset{} to include various sign language structures that use distinct channels (hands, face, or body) to convey information. As a case study, we have used \dataset{} to analyze the strengths and weaknesses of a state-of-the-art ASL-to-English translation model. We find that the model performs at above chance on almost all tested phenomena, and in cue ablation studies shows strong sensitivity to its hand input channel, but inconsistent sensitivity to non-manual channels, despite being trained on multi-channel inputs.  
Lastly, we have confirmed that standard machine translation evaluation (namely, BLEURT) cannot uncover the same detailed distinctions as minimal-pair analysis using \datasetshort{}. Overall, the minimal-pair analysis captures nuanced distinctions across linguistic structures and specific errors, helping pinpoint targeted improvements for future sign language models.  

We hope that our work will inspire additional linguistic studies of sign language models using \dataset{}, as well as additional minimal pair benchmarks, adding to the tradition of linguistic analyses of language models.  To the extent that more sign language translation models will be released publicly, \dataset{} will enable comparative studies across models.  Other potential directions for future work include building larger datasets, perhaps via automatic or semi-automatic discovery of phenomena-specific subsets in sign language video corpora, and extension to additional languages.

\section*{Acknowledgments}
Kanishka Misra is supported by the Donald D. Harrington Faculty Fellowship at UT Austin.

\bibliography{custom, kanishka, anthology, anthology_p2}
\bibliographystyle{acl_natbib}

\appendix

\section{Complementary Results}
\label{sec:complementary}

\begin{figure*}[!t]
    \centering
\includegraphics[width=\textwidth]{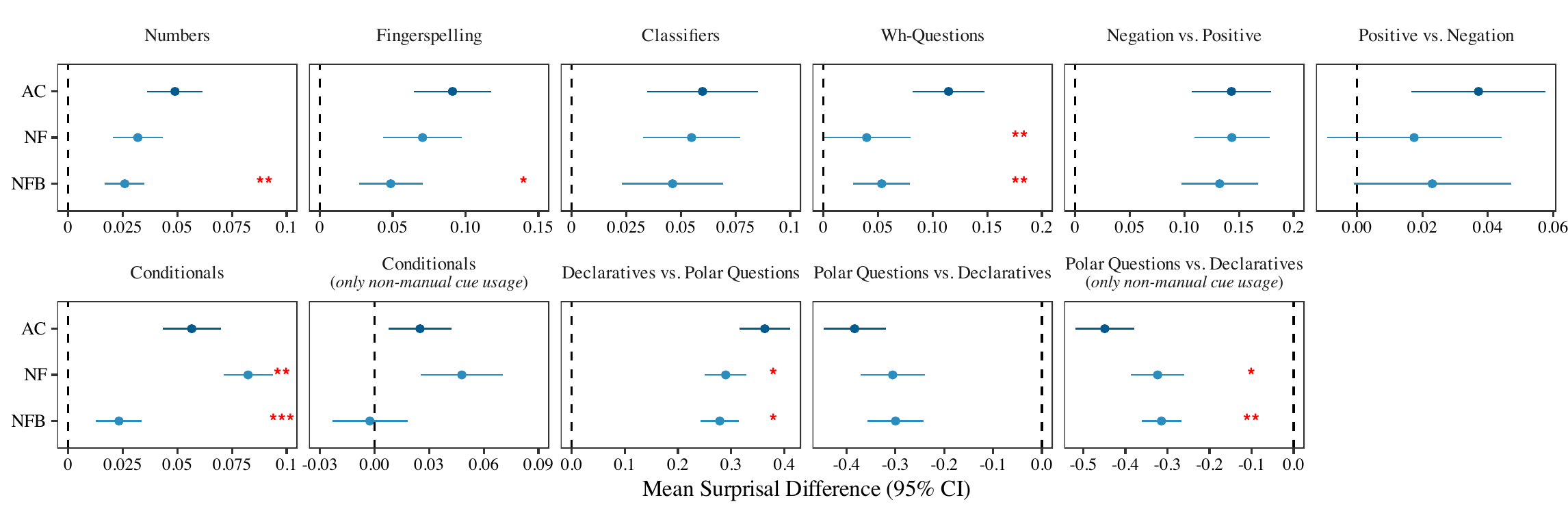}
    \caption{Average difference in surprisal of mismatched and matched sentences across phenomena and across \textbf{training ablations}. Error bars indicate 95\% confidence intervals. Stars (*) indicate significance test results for comparing surprisal difference in a given cue ablation to surprisal difference in the all cues condition (AC). *: \textit{p <} .05; **: \textit{p <} .01; ***: \textit{p <} .001}
    \label{fig:training-diff}
\end{figure*}

\Cref{fig:training-diff} shows the average surprisal differences ($\Delta\text{Surprisal}$) between the matched and the mismatched translations for the training time cue-ablations (i.e., ``controlled rearing'').

\clearpage

\end{document}